\documentclass[conference]{IEEEtran}
\ifCLASSINFOpdf
    \usepackage[pdftex]{graphicx}
\else
\fi
\usepackage{amsmath}
\usepackage{amsfonts}
\usepackage{algorithmic}
\usepackage[nolist]{acronym}
\usepackage{multirow}

\begin{document}
\IEEEoverridecommandlockouts

\title{A Novel Update Mechanism for Q-Networks Based On Extreme Learning Machines}

\author{\IEEEauthorblockN{Callum Wilson}
\IEEEauthorblockA{\textit{Department of Mechanical and Aerospace Engineering} \\
\textit{University of Strathclyde}\\
Glasgow, United Kingdom \\
callum.j.wilson@strath.ac.uk}
\and
\IEEEauthorblockN{Annalisa Riccardi}
\IEEEauthorblockA{\textit{Department of Mechanical and Aerospace Engineering} \\
\textit{University of Strathclyde}\\
Glasgow, United Kingdom \\
annalisa.riccardi@strath.ac.uk}
\and
\centering
\IEEEauthorblockN{Edmondo Minisci}
\IEEEauthorblockA{\textit{Department of Mechanical and Aerospace Engineering} \\
\textit{University of Strathclyde}\\
Glasgow, United Kingdom \\
edmondo.minisci@strath.ac.uk}%
\thanks{\copyright 2020 IEEE. Personal use of this material is permitted. Permission from IEEE must be obtained for all other uses, in any current or future media, including reprinting/republishing this material for advertising or promotional purposes, creating new collective works, for resale or redistribution to servers or lists, or reuse of any copyrighted component of this work in other works.}
}

\markboth{2020 IEEE World Conference on Computational Intelligence}%
{}
%


\maketitle
\begin{acronym}
\acro{AI}{Artificial Intelligence}
\acro{CSP}{Constraint Satisfaction Problem}
\acro{ELM}{Extreme Learning Machine}
\acro{EQLM}{Extreme Q-Learning Machine}
\acro{ESA}{European Space Agency}
\acro{LBR}{Low bit rate}
\acro{HBR}{High bit rate}
\acro{ML}{Machine Learning}
\acro{MSPA}{Multiple Spacecraft Per-Aperture}
\acro{NN}{Neural Network}
\acro{ROC}{Receiver Operating Characteristic}
\acro{RL}{Reinforcement Learning}
\end{acronym}

\begin{abstract}
    Reinforcement learning is a popular machine learning paradigm which can find near optimal solutions to complex problems. Most often, these procedures involve function approximation using neural networks with gradient based updates to optimise weights for the problem being considered. While this common approach generally works well, there are other update mechanisms which are largely unexplored in reinforcement learning. One such mechanism is Extreme Learning Machines. These were initially proposed to drastically improve the training speed of neural networks and have since seen many applications. Here we attempt to apply extreme learning machines to a reinforcement learning problem in the same manner as gradient based updates. This new algorithm is called Extreme Q-Learning Machine (EQLM). We compare its performance to a typical Q-Network on the cart-pole task - a benchmark reinforcement learning problem - and show EQLM has similar long-term learning performance to a Q-Network.
\end{abstract}


%
\IEEEpeerreviewmaketitle

\section{Introduction}
%
%
%
%

\label{sec:intro}

\IEEEPARstart{M}{achine} learning methods have developed significantly over many years and are now applied to increasingly practical and real world problems. For example, these techniques can optimise control tasks which are often carried out inefficiently by basic controllers. The field of \ac{RL} originates in part from the study of optimal control \cite{Bellman1954}, where a controller is designed to maximise, or minimise, a characteristic of a dynamical system over time. It is often impossible or impractical to derive an analytical optimal control solution for environments with complex or unknown dynamics, which motivates the use of more intelligent methods such as RL. In particular, intelligent controllers must be capable of learning quickly online to adapt to changes. The study of optimal control and RL brought machine learning into the broader field of engineering with applications to a wide variety of problems \cite{Sutton1998}.

The generalisation performance of \ac{RL}-derived controllers significantly improved with the incorporation of function approximators \cite{Sutton1996}. Unlike the earlier tabular methods, architectures such as fuzzy logic controllers \cite{Davoodi1998ComputerLearning} or more commonly \acp{NN} can exploit similarities in areas of the state space to learn better policies. This comes at a cost: \acp{NN} usually take a long time to train and in general they do not guarantee convergence. Furthermore, nonlinear function approximators can be unstable and cause the learning algorithm to diverge \cite{Tsitsiklis1997}. Despite this, through careful selection of hyperparameters and use of additional stability improvement measures, as will be discussed later, such function approximators can still obtain useful solutions to control problems. Of all the algorithms available for tuning network weights, backpropagation is the most widely used in state-of-the-art systems \cite{Silver2016,Mnih2015,Mnih2016,VanHasselt2016}. The most common alternatives to this approach involve evolutionary algorithms, which can be used to evolve network weights or replace the function approximator entirely \cite{Moriarty1999}. Such algorithms tend to show better performance but have a much higher computational cost which can make them infeasible for certain learning problems.

\acp{ELM} are a class of neural networks which avoid using gradient based updates \cite{Huang2006}. For certain machine learning problems, ELM has several advantages over other update rules - mainly that it can be considerably faster than iterative methods for optimising network weights since they are instead calculated analytically. ELM has seen many improvements and adaptations allowing it to be applied to a wide variety of problems involving function approximation \cite{Huang2011}. These include applications within the realm of RL, such as using a table to provide training data for an ELM network \cite{Lopez-Guede2013}, approximating system dynamics using ELM to later apply RL methods \cite{Lopez-Guede2014}, or using ELM theory to derive an analytical solution for weight updates based on the loss function of gradient-based updates \cite{Sun2015}. Here we aim to use ELM in a conventional RL algorithm by only altering the neural network update rule. The algorithm which uses ELM in this manner is referred to here as the ``Extreme Q-Learning Machine" (EQLM).

In this paper we develop the EQLM algorithm and compare its performance to a standard Q-network of the same complexity. The type of Q-network used here is relatively primitive but incorporates some features to improve stability and general learning performance to provide a reasonable comparison. A full stability analysis of each algorithm is outwith the scope of this paper, however we compare their performance using standard measures of learning. EQLM uses an incremental form of ELM which allows updates to be made online while the RL agent is interacting with the environment. Tests are carried out on a classical RL benchmark known as the cart-pole problem.

\section{Background}\label{sec:bkgd}

\begin{figure}[!t]
    \centering
    \includegraphics[width=0.4\textwidth]{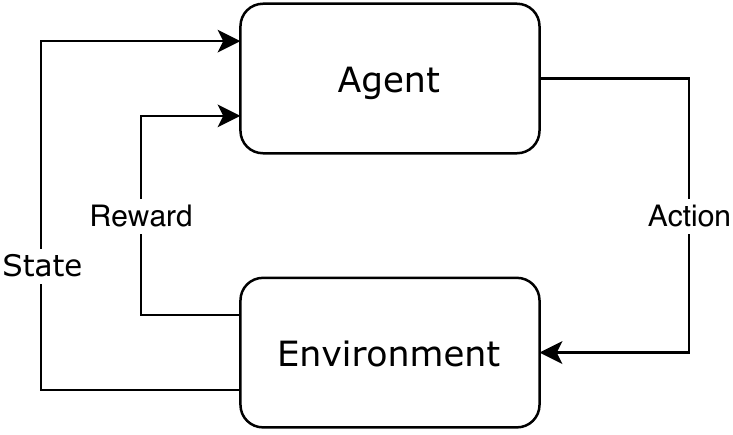}
    \caption{Agent-environment interaction in reinforcement learning, where the agent observes a state and reward signal from the environment and uses this information to select an action to take}
    \label{fig:agent_env}
\end{figure}

A RL process consists of an agent which senses an environment in a certain state and carries out actions to maximise future reward \cite{Sutton1998}. The only feedback the agent receives from the environment is a state signal and reward signal and it can only affect the environment by its actions as shown schematically in Figure \ref{fig:agent_env}. The objective is then to maximise the total discounted reward it receives. One method for optimising the reward is Q-Learning, where the agent learns the action-value function, $Q$ of its policy and uses this to improve the policy in an iterative process \cite{Watkins1989}. The temporal-difference (TD) error of $Q$ is defined as shown:

\begin{equation}
    e=Q(s_t,a_t)-\left ( r_{t+1}+\gamma \max_aQ(s_{t+1},a) \right )
    \label{eq:tderr}
\end{equation}

where $s_t$, $a_t$, and $r_t$ denote state, action, and reward respectively at time-step $t$, $\gamma$ is the discount factor which determines the affect of long term rewards on the TD error, and $Q(s,a)$ is the estimated action-value function. We approximate $Q(s,a)$ using a feed-forward NN with parameters $\theta$ and, for the standard Q-network, perform updates using the mean-squared TD error. Approximating the value function and updating using TD-error forms the basis for the most rudimentary Q-Learning algorithms. This section details the additional features of the Q-Learning algorithm which are employed in EQLM.

\subsection{{$\epsilon$}-Greedy Policies}

To find an optimal solution, an agent must visit every state in the environment throughout its training, which requires the agent to explore the environment by periodically taking random actions. This conflicts with the agent's global goal of taking actions deemed optimal by the current control policy to improve the policy; thus the well known issue of balancing exploration and exploitation. One type of policies which help remedy this issue are known as ``$\epsilon$-greedy" policies \cite{Sutton1998}. In these policies, a parameter $\epsilon$ dictates the probability of taking a random action at each time-step, where $0 \leq \epsilon \leq 1$, and this can be tuned to give the desired trade-off between taking exploratory or exploitative actions. Exploration becomes less necessary later in the training period once the agent has more experience. Instead, the agent seeks to exploit actions considered more ``optimal" following a period of more exploration. To achieve this in practice, $\epsilon$ varies linearly during the training period from $\epsilon_i$ to $\epsilon_f$ over $N_{\epsilon}$ episodes. Following this, $\epsilon$ is held constant at $\epsilon=\epsilon_f$ for the remainder of training. The exploration probability after $n$ episodes is given by Equation \ref{eq:eps}. 

\begin{equation}
    \epsilon=
    \begin{cases}
        \epsilon_i-\frac{n}{N_{\epsilon}}\left ( \epsilon_i - \epsilon_f \right ), & \text{if}\ n < N_{\epsilon}  \\
        \epsilon_f, & \text{if}\ n \geq N_{\epsilon}  \\
    \end{cases}
    \label{eq:eps}
\end{equation}

\subsection{Target Network}

A crucial issue with Q-networks is that they are inherently unstable and will tend to overestimate action-values, which can cause the predicted action-values to diverge \cite{Mnih}. Several methods to resolve this issue have been proposed, including the use of a target network \cite{Mnih2015}. This network calculates target action-values for updating the policy network and shares its structure with this network. The parameters of the policy network, $\theta$ are periodically transferred to the target network, whose parameters are denoted $\theta^-$, which otherwise remain constant. In practise, the target network values are updated every $C$ time-steps. This slightly decouples the target values from the policy network which reduces the risk of divergence.

\subsection{Experience Replay}

The online methods of learning discussed thus far all conventionally make updates on the most recent observed state transition, which has several limitations \cite{Mnih2015}. For example, states which are only visited once may contain useful update information which is quickly lost and updating on single state transitions results in low data efficiency of the agent's experience. A more data efficient method of performing updates utilises experience replay \cite{Lin1992}. In this method, experiences of transitions are stored in a memory, $\mathcal{D}$ which contains the state $s_j$, action taken $a_j$, observed reward $r_{j+1}$, and observed state $s_{j+1}$ for each transition. Updates are then made on a ``minbatch" of $k$ experiences selected randomly from the memory at every time-step. To limit the number of state transitions stored, a maximum memory size $N_{mem}$ is defined such that a moving window of $N_{mem}$ previous transitions are stored in the agent's memory \cite{Mnih}.

\subsection{Q-Network Algorithm}

Figure \ref{alg:qnet} details the Q-Network algorithm which incorporates a target network and experience replay. This algorithm gives our baseline performance to which we compare \ac{EQLM} and also provides the basis for incorporating \ac{ELM} as a novel update mechanism.

\begin{figure}
    \caption{Q-Network algorithm}
    \centering
    \begin{algorithmic}[1]
        \STATE initialise network with random weights
        \FOR{episode$=1$ to $N_{ep}$}
            \STATE initialise state $s_t \leftarrow s_0$
            \WHILE{state $s_t$ is non-terminal}
                \STATE select action $a_t$ according to policy $\pi$
                \STATE execute action $a_t$ and observe $r$, $s_{t+1}$
                \STATE update memory $\mathcal{D}$ with $\left ( s_t,a_t,r_t,s_{t+1} \right )$
                \STATE select random minibatch of k experiences $\left ( s_j,a_j,r_j,s_{j+1} \right )$ from $\mathcal{D}$
                \STATE $\mathbf{t}_{j}=
                \begin{cases}
                    r_j, & \text{if}\ s_{j+1} \text{ is terminal} \\
                    r_j+\gamma\max_{a}Q_T(s_{j+1},a), & \text{otherwise}
                \end{cases}$
                \STATE $e_j=Q(s_j,a_j)-\left ( r_{j+1}+\gamma \max_aQ_T(s_{j+1},a) \right )$
                \STATE update network using the error $e_j$ for each transition in the minibatch
                \STATE after $C$ time-steps set $\theta^{-} \leftarrow \theta$
            \ENDWHILE
        \ENDFOR
    \end{algorithmic}
    \label{alg:qnet}
\end{figure}

\section{ELM Theory and Development}\label{sec:elm}

\subsection{Extreme Learning Machine}

ELM in its most widely used form is a type of single-layer feedforward network (SLFN). The description of ELM herein uses the same notation as in \cite{Huang2006}. Considering an arbitrary set of training data $(\mathbf{x}_i,\mathbf{t}_i)$ where $\mathbf{x}_i=\left[ x_{i1},x_{i2},\dots,x_{in} \right]$ and $\mathbf{t}_i=\left[ t_{i1},t_{i2},\dots,t_{im} \right]$, a SLFN can be mathematically modelled as follows

\begin{equation}
    \sum_{i=1}^{\tilde{N}}\beta_ig\left( \mathbf{w}_i \cdot \mathbf{x}_j + b_i \right)=\mathbf{o}_j, \: \: j=1,\dots,N
\end{equation}
where $\tilde{N}$ is the number of hidden nodes, $\beta_i=\left[ \beta_{i1},\beta_{i2},\dots,\beta_{im} \right]^T$ is the output weight vector which connects the $i$th hidden node to the output nodes, $g(x)$ is the activation function, $w_i=\left[ w_{i1},w_{i2},...,w_{in} \right]^T$ is the input weight vector which connects the $i$th hidden node to the input nodes, and $b_i$ is the bias of the $i$th hidden node. Where the network output $\mathbf{o}_j$ has zero error compared to the targets $\mathbf{t}_j$ for all $N$ samples, $\sum_{j=1}^{\tilde{N}}\left \| \mathbf{o}_j -\mathbf{t}_j \right \|=0$ it can be written that

\begin{equation}
    \sum_{i=1}^{\tilde{N}}\beta_ig\left( \mathbf{w}_i \cdot \mathbf{x}_j + b_i \right)=\mathbf{t}_j, \: \: j=1,\dots,N
\end{equation}
which contains the assumption that the SLFN can approximate the $N$ samples with zero error. Writing this in a more compact form gives

\begin{equation}
    \mathbf{H}\beta=\mathbf{T}
    \label{eq:HBT}
\end{equation}
where $\mathbf{H}$ is the hidden layer output matrix, $\beta$ is the output weight vector matrix, and $\mathbf{T}$ is the target matrix. These are defined as shown

\begin{equation}
    \mathbf{H}=\begin{bmatrix}
    g(\mathbf{w}_1\cdot\mathbf{x}_1+b_1) & \cdots & g(\mathbf{w}_{\tilde{N}}\cdot\mathbf{x}_1+b_{\tilde{N}}) \\ 
    \vdots & \cdots & \vdots \\ 
    g(\mathbf{w}_1\cdot\mathbf{x}_N+b_1) & \cdots & g(\mathbf{w}_{\tilde{N}}\cdot\mathbf{x}_N+b_{\tilde{N}})
    \end{bmatrix}_{N\times\tilde{N}}
    \label{eq:H}
\end{equation}

\noindent\begin{minipage}{.5\linewidth}
\begin{equation}
    \beta=\begin{bmatrix}
    \beta_1^T \\
    \vdots \\
    \beta_{\tilde{N}}^T
    \end{bmatrix}_{\tilde{N}\times m}
    \label{eq:B}
\end{equation}
\end{minipage}%
\begin{minipage}{.5\linewidth}
\begin{equation}
    \mathbf{T}=\begin{bmatrix}
    \mathbf{t}_1^T \\
    \vdots \\
    \mathbf{t}_N^T
    \end{bmatrix}_{N\times m}
    \label{eq:T}
\end{equation}
\end{minipage}

ELM performs network updates by solving the linear system defined in equation \ref{eq:HBT} for $\beta$

\begin{equation}
    \hat{\beta}=\mathbf{H}^{\dagger}\mathbf{T}
    \label{eq:beta_hat}
\end{equation}
where $\mathbf{H}^{\dagger}$ here denotes the Moore-Penrose generalised inverse of $\mathbf{H}$ as defined in equation \ref{eq:pinv}. This is used since, in general, $\mathbf{H}$ is not a square matrix and so cannot be inverted directly. 

\begin{equation}
    \mathbf{H}^{\dagger}=\left( \mathbf{H}\mathbf{H}^T \right)^{\dagger} \mathbf{H}^T
    \label{eq:pinv}
\end{equation}

The method used by ELM to update its weights has several advantages over classical methods of updating neural networks. It is proven in \cite{Huang2006} that $\hat{\beta}$ is the smallest norm least squares solution for $\beta$ in the linear system defined by equation \ref{eq:HBT}, which is not always the solution reached using classical methods. ELM also avoids many of the issues commonly associated with neural networks such as converging to local minima and improper learning rate. Such problems are usually avoided by using more sophisticated algorithms, whereas ELM is far simpler than most conventional algorithms.

\subsection{Regularized ELM}

Despite the many benefits of ELM, several issues with the algorithm are noted in \cite{Deng2009}. Mainly, the algorithm still tends to overfit and is not robust to outliers in the input data. The authors propose a Regularized ELM which attempts to balance the empirical risk and structural risk to give better generalisation. This differs to the ELM algorithm which is solely based on empirical risk minimisation.

The main feature of regularized ELM is the introduction of a parameter $\bar{\gamma}$ which regulates the amount of empirical and structural risk. This parameter can be adjusted to balance the risks and obtain the best generalisation of the network. Weights are calculated as shown:

\begin{equation}
    \beta=\left( \frac{I}{\bar{\gamma}}+\mathbf{H}^TD^2\mathbf{H} \right)^{\dagger}\mathbf{H}^T\mathbf{T}
    \label{eq:RegELMLong}
\end{equation}

which incorporates the parameter $\bar{\gamma}$ and a weighting matrix $D$. Setting $D$ as the identity matrix $I$ yields an expression for unweighted regularized ELM:

\begin{equation}
    \beta=\left( \frac{I}{\bar{\gamma}}+\mathbf{H}^T\mathbf{H} \right)^{\dagger}\mathbf{H}^T\mathbf{T}
    \label{eq:RegELM}
\end{equation}

which is a simplification of equation \ref{eq:RegELMLong}. ELM is then the case of equation \ref{eq:RegELM} where $\bar{\gamma} \rightarrow \infty$. Adding the parameter $\bar{\gamma}$ adds some complexity to the ELM algorithm because of its tuning, however regularized ELM still maintains most of the advantages of ELM over conventional neural networks.

\subsection{Incremental Extreme Learning Machine}

It is desired to perform network updates sequentially on batches of data which necessitates an incremental form of ELM. Such an algorithm is presented in \cite{Guo2014} whose basis is the regularized form of ELM shown in equation \ref{eq:RegELM}. The algorithm used for the purposes of EQLM is the least square incremental extreme learning machine (LS-IELM). 

For an initial set of $N$ training samples $(\mathbf{x}_i, \mathbf{t}_i)$ the LS-IELM algorithm initialises the network weights as shown:

\begin{equation}
    \beta=A_t^{\dagger}\mathbf{H}^T\mathbf{T}
\end{equation}
where
\begin{equation}
    A_t=\frac{I}{\bar{\gamma}}+\mathbf{H}^T\mathbf{H}
\end{equation}
and $\mathbf{H}$ and $\mathbf{T}$ are given by equations \ref{eq:H} and \ref{eq:T}. Suppose new sets of training data arrive in chunks of $k$ samples - the hidden layer output matrix and targets for a new set of $k$ samples are as shown:

\begin{equation}
    \mathbf{H}_{IC}=\begin{bmatrix}
    g(\mathbf{w}_1\cdot\mathbf{x}_N+b_1) & \cdots & g(\mathbf{w}_{\tilde{N}}\cdot\mathbf{x}_N+b_{\tilde{N}}) \\ 
    \vdots & \cdots & \vdots \\ 
    g(\mathbf{w}_1\cdot\mathbf{x}_{N+k}+b_1) & \cdots & g(\mathbf{w}_{\tilde{N}}\cdot\mathbf{x}_{N+k}+b_{\tilde{N}})
    \end{bmatrix}_{k\times\tilde{N}}
    \label{eq:H_IC}
\end{equation}

\begin{equation}
    \mathbf{T}_{IC}=\begin{bmatrix}
    \mathbf{t}_{N+1}^T \\
    \vdots \\
    \mathbf{t}_{N+k}^T
    \end{bmatrix}_{k\times m}
    \label{eq:T_IC}
\end{equation}

To perform updates using the most recent data at time $t$, $K_t$ is defined as
\begin{equation}
    K_t=I-A_t^{\dagger}\mathbf{H}_{IC}^T\left( \mathbf{H}_{IC}A_t^{\dagger}\mathbf{H}_{IC}^T + I_{k\times k} \right)^{\dagger} \mathbf{H}_{IC}
    \label{eq:kt_update}
\end{equation}
and the update rules for $\beta$ and $A$ are then as follows:
\begin{equation}
    \beta_{t+1}=K_t\beta_t+K_tA_t^{\dagger}\mathbf{H}_{IC}^T \mathbf{T}_{IC}
    \label{eq:beta_update}
\end{equation}
\begin{equation}
    A_{t+1}^{\dagger}=K_tA_t^{\dagger}
    \label{eq:at_update}
\end{equation}

\subsection{Extreme Q-Learning Machine}

The algorithm for applying Q-learning using LS-IELM based updates, here referred to as the Extreme Q-Learning Machine (EQLM) is shown in Figure \ref{alg:eqlm}. Similar to the Q-network algorithm in Figure \ref{alg:qnet}, this uses experience replay and a target network to improve its performance. Unlike the Q-network, the TD-error is not calculated and instead a target matrix, $\mathbf{T}$ for the minibatch of data is created which has the predicted action-values for all actions in the given states. The target action-value for each state, $s_j$ is then assigned to the applicable value in $\mathbf{t}_j$. Matrix $\mathbf{H}$ is constructed using the states in the minibatch and then the update rules are applied. The boolean variable $step0$ is introduced to initialise the network at the very first update.

One further key difference in the EQLM algorithm is the heuristic policy used in initial episodes. The return in initial episodes has a substantial effect on the convergence of EQLM as discussed later. This necessitates a simple heuristic controller for the start of training which does not need to perform very well, but can at least prevent the agent from converging on a highly sub-optimal policy. EQLM uses a heuristic action selection $a_t=h_0(t)$, which is effectively an open loop control scheme dependant only on the time-step, for $N_h$ episodes. Definition of this heuristic is discussed in the following section.

\begin{figure}[!t]
    \caption{EQLM algorithm}
    \centering
    \begin{algorithmic}[1]
        \STATE initialise network with random weights
        \STATE $step0\leftarrow True$
        \FOR{episode$=1$ to $N_{ep}$}
            \STATE initialise state $s_t \leftarrow s_0$
            \WHILE{state $s_t$ is non-terminal}
                \IF{episode$\leq N_h$}
                    \STATE select action $a_t$ according to heuristic $h_0(t)$
                \ELSE
                    \STATE select action $a_t$ according to policy $\pi$
                \ENDIF
                \STATE execute action $a_t$ and observe $r$, $s_{t+1}$
                \STATE update memory $\mathcal{D}$ with $\left ( s_t,a_t,r_t,s_{t+1} \right )$
                \STATE select random minibatch of k experiences $\left ( s_j,a_j,r_j,s_{j+1} \right )$ from $\mathcal{D}$
                \STATE $\mathbf{t}_j=
                \begin{cases}
                    r_j, & \text{if}\ s_{j+1} \text{ is terminal} \\
                    r_j+\gamma\max_{a}Q(s_{j+1},a), & \text{otherwise}
                \end{cases}$
                \STATE construct matrix $\mathbf{H}$
                \IF{$step0$}
                    \STATE $A_t=\frac{I}{\bar{\gamma}}+\mathbf{H}^T\mathbf{H}$
                    \STATE $\beta_{t+1}=A_t^{\dagger}\mathbf{H}^T\mathbf{T}$
                    \STATE $A_{t+1}=A_t$
                    \STATE $step0\leftarrow False$
                \ELSE
                    \STATE $K_t=I-A_t^{\dagger}\mathbf{H}^T\left( \mathbf{H}A_t^{\dagger}\mathbf{H}^T + I_{k\times k} \right)^{\dagger} \mathbf{H}$
                    \STATE $\beta_{t+1}=K_t\beta_t+K_tA_t^{\dagger}\mathbf{H}^T \mathbf{T}$
                    \STATE $A_{t+1}^{\dagger}=K_tA_t^{\dagger}$
                \ENDIF
                \STATE after $C$ time-steps set $\theta^{-} \leftarrow \theta$
            \ENDWHILE
        \ENDFOR
    \end{algorithmic}
    \label{alg:eqlm}
\end{figure}

\section{Experiments and Results}

Code to reproduce results is available at https://github.com/strath-ace/smart-ml. 

\subsection{OpenAI Gym Environments}
The environment used to test the algorithms comes from the OpenAI Gym which is a toolkit containing benchmark tests for a variety of machine learning algorithms \cite{Brockman2016}. The gym contains, among other environments, several classical control problems, control tasks which use the MuJoCo physics engine \cite{Todorov2012}, and the Atari2600 games which are used in \cite{Mnih2015}. Here the agents will be tested on the environment named ``CartPole-v0".

The cart-pole problem was originally devised in \cite{Michie1968} where the authors created an algorithm called ``BOXES" to learn to control the system. In this task, a pole is attached by a hinge to a cart which rolls along a track and is controlled by two possible actions - an applied force of fixed magnitude in either the positive or negative x-direction along the track. The goal is to keep the pendulum from toppling for as long as possible, which yields a very simple reward function of $r=+1$ for every time-step where the pendulum has not toppled. In addition, the track on which the cart is situated is finite and reaching the limits of the track also indicates failure. The dynamics of the system used in the gym are the same as those defined by \cite{Barto1983}. The state-space size for this environment is 4 and the action-space size is 2.

This problem can be considered an ``episodic" task \cite{Sutton1998}, where the learning is divided into episodes which have defined ending criteria. An episode terminates either when the pendulum passes $12^{\circ}$ or the cart reaches either end of the track. In this task, a maximum number of time-steps per episode of 200 is set within the gym.

\subsection{Heuristic Policy}
\begin{figure}[ht]
    \centering
    \includegraphics[width=0.48\textwidth]{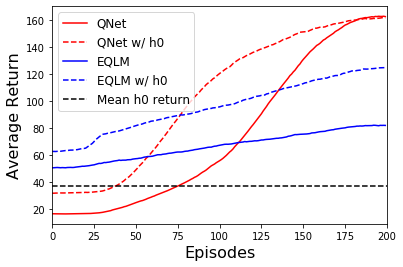}
    \caption{Varying performance with the use of an initial heuristic $h_0$ with the average performance for the heuristic alone shown}
    \label{fig:heur}
\end{figure}

As discussed previously, EQLM is susceptible to converging on a sub-optimal policy without the use of a heuristic policy in the initial episodes. A random policy at the start of training will sometimes produce this sub-optimal result and so we need to define a simple deterministic policy which does not immediately solve the task but prevents unacceptable long-term performance. For the cart-pole task we consider here which has a binary action space, we define the heuristic policy as taking alternating actions at each time-step as shown:
\begin{equation}
    h_0(t)=mod(t,2)
\end{equation}
From testing, we found $N_h=5$ to be a suitable number of episodes over which to use the heuristic. The effect of this initial heuristic policy is shown in Figure \ref{fig:heur}. This shows the averaged rewards over the first 200 episodes of training for both networks with and without the heuristic. While the return in the initial episodes is still higher for EQLM in both cases, it is clear that with the heuristic EQLM shows a more favourable performance. This is due to occasions where, without the heuristic, EQLM quickly converges to a sub-optimal policy, which is mitigated by the heuristic policy. Also shown is the average performance of the heuristic alone, which receives a reward of 37 per episode. This indicates that although the heuristic alone performs very poorly on the task, it is still useful to improve the performance of both algorithms.

\subsection{Hyperparameter Selection}
The performance of a Q-learning agent can be very sensitive to its hyperparameters. To create a useful comparison of each agent's performance we therefore need to tune the hyperparameters for this problem. Here we use the Python library Hyperopt which is suitable for optimising within combined discrete- and real-valued search spaces \cite{Bergstra2013}. Hyperparameters to be optimised are: learning rate $\alpha$ (Q-Network only), regularisation parameter $\bar{\gamma}$ (EQLM only), number of hidden nodes $\Tilde{N}$, initial exploration probability $\epsilon_i$ (with $\epsilon_f$ fixed as 0), number of episodes to decrease exploration probability $N_\epsilon$, discount factor $\gamma$, minibatch size $k$, and target network update steps $C$.

\begin{figure}[ht]
    \centering
    \includegraphics[width=0.48\textwidth]{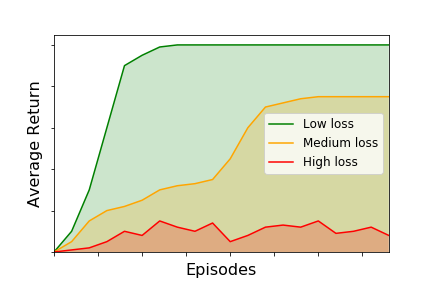}
    \caption{Example learning curves which show different values for the loss function}
    \label{fig:loss}
\end{figure}

Our main objective to optimise is the final performance of the agent, i.e. the total reward per episode, after it converges to a solution. In addition, an agent should converge to the optimal solution in as few episodes as possible. Both these objectives can be combined into the single metric of area under the learning curve as shown in Figure \ref{fig:loss}. Since hyperopt uses a minimisation procedure, we specifically take the negative area under the curve. One of the issues with optimising these systems is their stochastic nature which can result in several runs with the same hyperparameters having vastly different performance. To account for this, each evaluation uses 8 runs and the loss is the upper $95\%$ confidence interval of the metric from these runs. This gives a conservative estimate of the worst-case performance for a set of hyperparameters.

Table \ref{tab:params_cartpole} shows the best parameters obtained when tuning the hyperparameters for this task. Most of the hyperparameters common to each algorithm are not substantially different with the exception of minibatch size, $k$ which is 26 and 2 for the Q-network and EQLM respectively. In fact, the performance of EQLM tended to decrease for larger values of $k$ which was not the case for the Q-network. This could be a result of the matrix inversion in EQLM where the behaviour of the network is less stable if the matrix is non-square. Alternatively, it is possible that EQLM attempting to fit to a much larger number of predicted Q-values causes the overall performance to decrease. The fact it needs fewer data per time-step than a standard Q-network could also indicate that EQLM is more efficient at extracting information on the environment's action-values compared to using gradient descent.
\begin{table}[ht]
    \centering
    \begin{tabular}{|c|c|c|}
    \hline
        Hyperparameter & Q-Network  & EQLM  \\ \hline
        $\alpha$        & 0.0065    & - \\
        $\bar{\gamma}$  & -         & 1.827e-5 \\
        $\Tilde{N}$     & 29        & 25 \\
        $\epsilon_i$    & 0.670     & 0.559 \\
        $N_{\epsilon}$  & 400       & 360 \\
        $\gamma$        & 0.99      & 0.93 \\
        $k$             & 26        & 2 \\
        $C$             & 70        & 48 \\ \hline
    \end{tabular}
    \vspace{5pt}
    \caption{Hyperparameters used for each agent in the cart-pole task}
    \label{tab:params_cartpole}
\end{table}

\subsection{Learning Performance}

\begin{figure}[ht]
    \centering
    \includegraphics[width=0.48\textwidth]{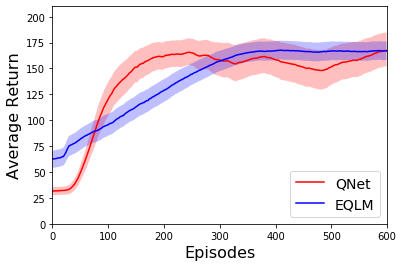}
    \caption{Learning curves for EQLM and a standard Q-Network in the cart-pole task. Results are averaged over all 50 runs at each episode and shaded area indicates the $95\%$ confidence interval}
    \label{fig:learn_compare}
\end{figure}

With the selected hyperparameters, each agent carried out 50 runs of 600 episodes in the cart-pole environment to compare their performance. The results of this are shown in Figure \ref{fig:learn_compare} and Table \ref{tab:results_cartpole}. Here we use two measures of performance: mean reward over the final 100 episodes and area under the learning curve (auc).

\begin{table}[ht]
    \centering
    \begin{tabular}{|c|c|c|c|}
    \hline
        \multicolumn{2}{|c|}{Measure}      & Q-Network & EQLM  \\ \hline
        \multirow{2}{*}{reward}     & mean & 160.0 (147.5, 173.7) & 166.9 (160.7, 173.3) \\
                                    & std  & 47.0 (35.1, 62.2) & 23.1 (20.3, 26.7) \\ \hline
        \multirow{2}{*}{auc ($*10^3$)} & mean & 84.1 (81.0 87.4) & 83.3 (80.4, 86.2) \\
                                    & std  & 11.7 (9.1, 14.7) & 10.6 (9.3, 12.4) \\ \hline
    \end{tabular}
    \vspace{5pt}
    \caption{Performance of each algorithm in the cart-pole task}
    \label{tab:results_cartpole}
\end{table}

From the learning curves, we see EQLM on average achieves a superior performance in the earliest episodes of training followed by a steady increase in return until it plateaus. The Q-network begins with comparatively low average return but then shows a sharp increase in return before its performance plateaus for the remainder of the episodes. After each of the learning curves plateau at their near-optimal performance, we see some of the most interesting differences between the two algorithms. The average return for EQLM remains very consistent as do the confidence intervals, however the Q-network displays some temporal variation in its performance as training continues and the confidence intervals tend to get larger. This shows that the long-term performance of EQLM is more consistent than the equivalent Q-network, which is backed up by the data in Table \ref{tab:results_cartpole}. The standard deviation of the mean reward of EQLM (23.1) is less than half that of the Q-network (47.0) and both algorithms have comparable mean rewards (160.0 and 166.9 for Q-network and EQLM respectively).

To find a statistical measure of the difference in performance of each algorithm, we use a two-tailed t-test \cite{Henderson2018}. This assumes both algorithms' performance belongs to the same distribution which we reject when the p-value is less than a threshold of 0.05. When comparing the mean reward in the final episodes, the t-test yielded values of $t=-0.628$, $p=0.531$. Similarly for the area under the learning curve we obtained $t=-1.16$, $p=0.24$. As a result, we cannot reject the hypothesis that the performance of both algorithms follows the same distribution. This demonstrates EQLM as being capable of achieving similar performance to a standard Q-Network in this task.

\section{Conclusion}
This paper proposed a new method of updating Q-networks using techniques derived from ELM called Extreme Q-Learning Machine (EQLM). When compared to a standard Q-network on the benchmark cart-pole task, EQLM shows comparable average performance which it achieves more consistently than the Q-network. EQLM also shows better initial learning performance when initialised using a basic heuristic policy.

While EQLM shows several advantages to standard Q-networks, it is clear that the conventional gradient descent methods are also capable of learning quickly as they gain more experience. Future work could look at combining the strengths of EQLM's initial performance and using gradient-based methods to accelerate the learning. In this paper we have tuned the hyperparameters of EQLM for a specific problem, but a more rigorous parametric study is necessary to learn more about the effect of the hyperparameters on EQLM's learning performance. One of the developments in ELM which was not used here is the ELM-based multilayer perceptron \cite{Tang2015}. Such a network could similarly be used for RL problems since deep networks are generally better suited to more complex tasks \cite{LeCun2015}. 

The results in this paper suggest ELM methods are capable of being used within RL with similar performance and greater consistency than conventional gradient-descent for simple RL problems. Additional research is needed on the application of EQLM to higher dimensional and adaptive control problems.


%



\section*{Acknowledgment}
The authors would like to thank the University of Strathclyde Alumni Fund for their support.


\ifCLASSOPTIONcaptionsoff
  \newpage
\fi



\bibliographystyle{IEEEtran}
\bibliography{IEEEabrv,references_arp.bib}

\end{document}